# Which Factors Predict the Chat Experience of a Natural Language Generation Dialogue Service?

Eason Chen
National Taiwan Normal University
Program of Learning Sciences
Taipei, Taiwan
eason.tw.chen@gmail.com

## ABSTRACT

In this paper, we proposed a conceptual model to predict the chat experience in a natural language generation dialog system. We evaluated the model with 120 participants with Partial Least Squares Structural Equation Modeling (PLS-SEM) and obtained an R-square ($R^2$) with 0.541. The model considers various factors, including the prompts used for generation; coherence, sentiment, and similarity in the conversation; and users' perceived dialog agents' favorability. We then further explore the effectiveness of the subset of our proposed model. The results showed that users' favorability and coherence, sentiment, and similarity in the dialogue are positive predictors of users' chat experience. Moreover, we found users may prefer dialog agents with characteristics of Extroversion, Openness, Conscientiousness, Agreeableness, and Non-Neuroticism. Through our research, an adaptive dialog system might use collected data to infer factors in our model, predict the chat experience for users through these factors, and optimize it by adjusting prompts.

## CCS CONCEPTS

• **Human-centered computing** → User models; **HCI theory, concepts and models**; • **Computing methodologies** → **Natural language generation**; **Discourse, dialogue and pragmatics**.

## KEYWORDS

Chatbot, Dialogue System, User Experience, Coherence, Big Five Personality, Partial Least Squares Structural Equation Modeling

## 1 INTRODUCTION

A chatbot is an application where users can interact through natural language in a conversational interface [13]. In recent years, chatbot for open-domain dialog has become popular thanks to advances in the Natural Language Generation (NLG) technique [1, 11, 16]. Researchers interested in building NLG chatbot and conducted much research in enhancing the chat experience for chatbot users. These research include changing the chat agent's avatar [5, 6], using multiple agents [2, 3], and setting proper agent character [19]. Previous research also indicated that messages' coherence [7], sentiment [14] are related to a good conversation experience. Nevertheless, as far as we know, no research investigated the relationship between these factors. This is the reason we wrote this paper. There is huge potential in analyzing these factors and using them to predict users' chat experience. By doing so, an adaptive dialog system might use the collected data to infer these factors and to predict or even optimize users' chat experience.

In this paper, our research questions are 'which factors predict the Chatting Experience of a Natural Language Generation Dialogue Service?' and 'how are these factors related to each other?' To answer these questions, we summarize factors that will influence users' chatting experience in an NLG dialog service. Then we conducted an experiment with 120 participants to collect users' feedback and preferences on NLG dialog agents with different configurations. Finally, based on the collected data, we analyze the relationship between these factors using Partial Least Squares Structural Equation Modeling (PLS-SEM). In sum, the main contributions of this paper are:

(1) We proposed a conceptual model to predict the chat experience in natural language generation dialog systems which obtained an R-square ($R^2$) with 0.541 in PLS-SEM.
(2) We find that users' favorability on agents and coherence, sentiment, and similarity in the conversation are strong positive predictors to chat experience.
(3) Our results indicated that users may prefer dialog agents with characteristics of Extroversion, Non-Neuroticism, Openness, Conscientiousness, and Agreeableness.

## 2 THEORETICAL FRAMEWORK AND HYPOTHESES

### 2.1 Natural Language Generation and Chat Experience

Natural language generation (NLG) uses artificial intelligence to produce natural language in a way that resembles human language [16]. NLG is done by analyzing the given input and using it to generate grammatically correct text that accurately reflects the input [11]. NLG is widely used in the dialog system nowadays. The models and techniques used to generate dialog agents' natural language replies are constantly improving.

Thanks to large NLG models like GPT-3 [1, 16], the generation process can now fine-tune the generated output by prompts. For example, given the same input in line 2: "The boss keeps making things difficult for me", chat agents (Friend) in examples 1 and 2 with different character configurations by prompts in line 1, will provide completely different replies in line 3 and line 5 shown in bold text.

(Example 1)
1. The following is your conversation with your friend, who is down-to-earth, insensitive, and conventional.
2. You: The boss keeps making things difficult for me.

3. Friend: **That's just the way the world works. You can't always get what you want.**

(Example 2)
1. The following is your conversation with your friend, who is good-natured, compliant, modest, and cooperative.
2. You: The boss keeps making things difficult for me.
3. Friend: **I'm sorry to hear that. What can I do to help?**
4. You: Encourage me.
5. Friend: **You definitely have the ability to change the situation. You may need to speak with your boss directly about your concerns, or find a way to work more efficiently. Stay positive and don't give up!**

Furthermore, we can also perform multi-turn conversations with GPT-3, which takes previous messages as prompts with the input as the dialog context. For example, in the follow-up conversation in example 2, given the previous conversation and the input 'encourage me', the reply (bold text) from the agent (Friend) can refer to 'boss' in the previous conversation.

Based on these advanced NLG models, previous research on enhancing and evaluating the quality of the generated output focus on improving messages' coherence [7], sentiment [14], and emotion [20, 21]. Specifically, previous research concluded that dialog with higher coherence [7] and sentiment score [14] will lead to a more satisfiable conversation. Therefore, we suppose that dialogs' coherence and sentiment can predict users' chat experience and their favorability score on agents:

- H1a Messages' coherence and sentiment can predict users' chat experience
- H1b Messages' coherence and sentiment can predict users' favorability on agents

## 2.2 Dialog Agent and Chat Experience

A chatbot is an application designed to simulate a conversation between a human user and a virtual agent. As the target of interaction, the design of the chat agent is very critical. Previous research indicated that the appearance [6], dialog context, and character design [19] of the chat agent would influence users' attitudes toward them. Moreover, as research indicated that first impressions would greatly influence people's opinions [20], we will measure both users' first impressions and follow-up favorability scores on agents. The former might be able to evaluate agents' appearance only, while the latter users might also take chat agents' character design and dialog context into account.

- H2a Users' pre-test favorability score can predict their post-test favorability score
- H2b Users' pre-test favorability score and follow-up favorability score can predict their chat experience

Moreover, this research will explore how different chat agents' characters influence users' chat experience and favorability. We will use prompt programming with Big Five personality as the basis for agents' character design. The Big Five personality [18] is a set of factors that describe most personality traits: Neuroticism,

Openness To Experience, Extraversion, Agreeableness, and Conscientiousness. We also interested in how these different prompts design will moderate users' chat experience.

- H3 Different prompts will lead to different chat experience

To let users compare, we will present five replies from five dialog agents with different character designs at once. Previous research indicated that a dialog system with multiple agents would bring a better chat experience because users can pay attention to the reply they like [4]. Nevertheless, another study [2] also stated that users would feel tiresome if multiple replies were similar. Therefore, in this paper, we suppose that the similarity of replies from different agents would predict users' favorability and chat experience.

- H4a The similarity between each agent's replies can predict users' chat experience
- H4b The similarity between each agent's replies can predict users' favorability score

## 2.3 Proposed Model

Based on the above theoretical discussion, we propose the following conceptual model in Figure 1. In our experiment, we will not only investigate the model with partial least squares structural equation modeling but also compare the effect of different groups of users and chatbot design.

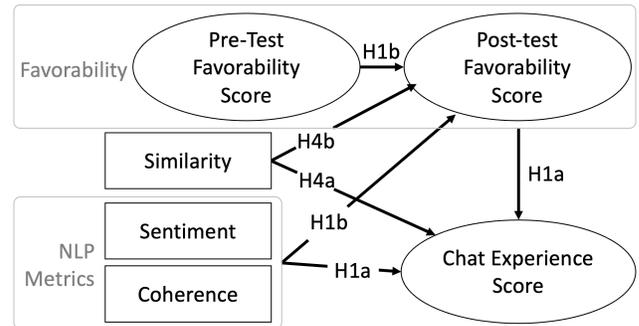

**Figure 1: The proposed conceptual model.**

## 3 RESEARCH METHODS

### 3.1 Participants

We recruited 120 participants in Taiwan from the internet through convenience sampling with a mean age of 24.7 years (range 18 − 64, SD = 7.9).

### 3.2 Experiment Design and Procedure

The goal of our experiment is to let users evaluate their favorability and chat experience on different chat agents in the conversation. To do so, we create 20 chat agents with different personalities, 5 (Big Five, ex: Extraversion) x 2 (opposite of Big Five, ex: Introversion) x 2 (Gender). We set agents with Big5 personality prompts description suggested by [18]. For example, if we want a female chatbot named Liz who is Agreeableness, we can use the prompt: 'Your friend is a woman named Liz. The following is your conversation with your friend, who is good-natured, compliant, modest, gentle, and cooperative.' Additionally, the names of the bots were randomly assigned, and the avatars were carefully chosen from the AffectNet dataset [15]. The avatars were selected with a happy facial expression label, natural image style, and corresponding gender. Examples of the inputs and agent configurations can be found in section 2.1 and in the supplemental material.

Our research is approved by the Institutional Review Board. The experiment was divided into four rounds of chats, with 20 bots split into 4 groups of 5 bots each. Participants began by rating their attachment style and then engaged in four rounds of conversation in random order. During each round, participants rated their favorability of each bot based on its name and avatar. They then engaged in ten conversation turns with the five bots in their group before rating their favorability, estimated personality, and the overall quality of the chatting experience for each bot.

The experiment was conducted on LINE, a widely-used instant messaging service in Taiwan. To achieve this, we created a LINE bot that could change both its avatar and the name of its responses. Additionally, we utilized the LINE Front-end Framework to gather user feedback during each experimental condition. This allowed participants to complete the experiment using only the LINE app and a mobile web browser. The source code for the platform can be found at GitHub[1].

### 3.3 Natural Language Generation Model Design

We utilize the text-davinci-001, a GPT-3 model, with English as the primary language as the NLG model of dialog agents. The prompts used for generated text include the bot's personality and previous three dialogs for multi-turn conversation. Examples of input and bot settings can be found in Section 2.1 and Appendix A in Supplementary Materials.

### 3.4 Data Collection

*3.4.1 Chat Experience.* We translated the Chatbot Usability Questionnaire (CUQ) [10] into Chinese to measure users' chat experience toward each agent. The purpose of the CUQ is to assess the participant's perception of their experience using chatbots. To do this, it presents 16 statements, such as 'The chatbot understood me well' or 'Chatbot responses were irrelevant', which the participant must rate on a scale of 1 (strongly disagree) to 5 (strongly agree). The Cronbach's Alpha of the collected CUQ data in the second experiment is $\alpha = .95$. Please see appendix B for all CUQ questions we used.

*3.4.2 Favorability.* Users will first rate their favorability on a scale of 1 to 7 based on the name and avatar of the agent. After finished the chat, they will re-rate the favorability score of the agent based on the name, avatar, and chat history.

*3.4.3 Coherence.* We use the microsoft/DialogRPT-human-vs-rand model [7] to evaluate the coherence of a dialog. Given the input and reply, the model can predict a score between 0 and 1 that represents the likelihood of the response being relevant to the input. We evaluate both user input and bot reply, as well as the bot reply and follow-up input from users. During analysis, we will calculate the average coherence score for each agent and user.

*3.4.4 Sentiment.* We use the distilbert-base-uncased-finetuned-sst-2-english model to predict the sentiment of the dialog. The model, which is widely used for sentiment analysis at Hugging Face, outputs a positive score for a given text on a scale of 0 to 1. When analyzing, we calculated the average score of users' inputs and agents' replies.

*3.4.5 Similarity.* We use the sentence-BERT [17] to perform embedding on outputs with the same input but with different prompts. Then we calculate the mean of the cosine similarity on these outputs to determine the similarity score on each generated output.

### 3.5 Data Analysis

We collected 2400 samples (120 participants x 20 agents). We used SmartPLS 4 to perform the Partial Least Squares Structural Equation Modeling (PLS-SEM). PLS-SEM is a second-generation multivariate analysis technique useful for examining relationships between latent independent and dependent variables [8].

We first used PLS-SEM to verify the construct validity of the proposed model. We evaluated the reliability, internal consistency, and convergent and discriminant validity through the confirmatory factor analysis. All latent variables showed good reliability and internal consistency after eliminating some questions. Due to the word limitations, detailed results of the reliability and validity are presented in the Supplemental Material.

Then, we used PLS-SEM to examine structural relationships among latent variables on the proposed model. The maximum likelihood method and bootstrapping resampling were employed to measure the statistical significance of the model's path coefficients.

Furthermore, we also examine the effectiveness by R-square with parts of the factors by performing PLS-SEM with the subset of the proposed model. Especially, we remove the posttest favorability score because it might yield similar nature to CUQ.

Finally, we evaluate the chat experience score of different chat agents designed by ANOVA, then divide the agents into two subgroups. By examining models from subgroups, we identified differences between dialog agents with high and low chat experience scores.

## 4 RESULTS

### 4.1 Construct validity of the proposed model

The adjusted R square ($_{adj}R^2$) of the chat experience score on our proposed model is 0.541. All the item factor loadings are above 0.65. The average variance extracted (AVE) of all latent variables are above 0.5. The composite reliability (CR) and Cronbach's alpha (CA) of all latent variables are above 0.7. Moreover, the Standardized Root Mean Square Residual (SRMR) of the estimated model is 0.06, which means the model fit is very good [12].

### 4.2 Structural relationships among latent variables

The PLS-SEM was used to examine the model of the structural relationships among the latent variables of the proposed model.

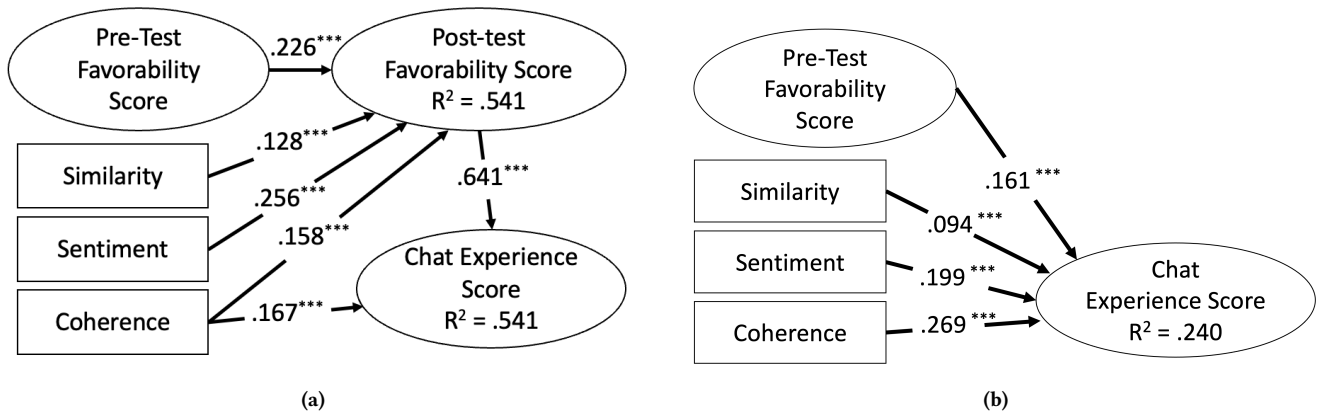

**Figure 2: The PLS-SEM result of (a) the proposed model. (b) the proposed model without post-test favorability score.**

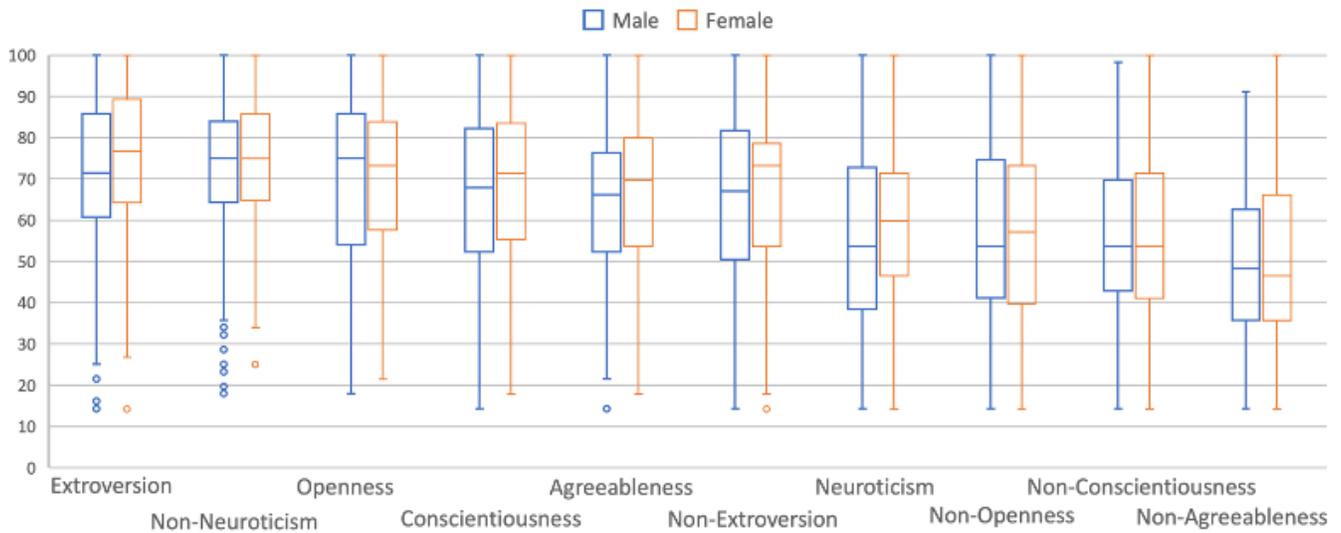

**Figure 3: The difference of chat experience score between different agents' Big Five personality and gender**

The result is shown in Figure 2. Paths with no statistical significance were ignored. Because of the large sample size (N = 2400), we consider $p < .001$ as significant and $p < .005$ as marginally significant.

The result (in Figure 2) indicated that the chat experience score ($_{adj}R^2 = .541$) can be predicted by post-test favorability score (path coefficient = .641, $p < .001$) and coherence (path coefficient = .167, $p < .001$). While the post-test favorability score (after chat) ($_{adj}R^2$ = .25) can be predicted by first impression (path coefficient = .226, $p < .001$), coherence (path coefficient = .158, $p < .001$), similarity (path coefficient = .128, $p < .001$), and sentiment (path coefficient = .256, $p < .001$).

### 4.3 Examine the effectiveness on with parts of the factors

In this section, we investigate the value of adjusted R-square ($_{adj}R^2$) with parts of the factors (latent variables). If we remove the post-test favorability score (Figure 2.b), the chat experience score ($_{adj}R^2$ = 0.240) can be predicted by pre-test favorability score (path coefficient = .161, $p < .001$) and coherence (path coefficient = .269, $p < .001$), sentiment (path coefficient = .199, $p < .001$), and similarity (path coefficient = .094, $p < .001$) in the conversation.

If we keep the post-test favorability score only, the $_{adj}R^2$ will be 0.51. This mean the post-test favorability score is a very strong predictor and mediator. Nevertheless, the Discriminant Validity Assessment and Heterotrait-monotrait Ratio of Correlations (HTMT) between the posttest favorability score and the chat experience score is 0.728, indicating they belong to two different factors (HTMT should < 0.9) [9].

## 4.4 Comparison between agents with high and low chat experience score

A two-way ANOVA was conducted on chat experience score for 20 conditions: 10 (Big Five personality and their opposites) x 2 (genders). The interaction between personality and gender was not significant ($F(9, 2380) = .288$, $p = .978$). Moreover, the main effects analysis revealed that agents' prompts designed by Big Five personality ($F = 42.2$, $p < .001$, $\eta^2 = .138$) had a significant effect on chat experience. (H3 accepted). The results in Figure 3 indicated that agents with Extroversion, Non-Neuroticism, Openness, Conscientiousness, and Agreeableness personalities will have higher chat experience scores ($t(2398) = 15.96$, $p < .001$).

Then we compared the difference in paths of the proposed conceptual model between agents with prompts that will lead to higher and lower chat experience scores. We found that for higher chat experience agents, similarity is no longer a significant predictor.

## 5 DISCUSSION

### 5.1 Effectiveness of the proposed model

Our conceptual model was found to fit well and strongly predict chat experience scores ($_{adj}R^2 = .541$). Additionally, it remained a strong predictor even when using only a subset of the model, such as the favorability score ($_{adj}R^2 = .51$) or first impression and natural language model metrics ($_{adj}R^2 = .24$), including coherence, sentiment, emotion, and similarity. Therefore, the researcher and developer might use factors in our model to predict and optimize users' chat experience in an adaptive dialog system.

### 5.2 The importance of users' favorability on agents

Our findings showed that the favorability score strongly predicts the chat experience. Therefore, we suggest that simply asking users how they feel about the dialog agents is an effective way to estimate users' chat experience.

Additionally, while the agents' pre-test favorability score was related to the chat experience score, its' effect was mitigated by post-test favorability. The key takeaway is that while first impressions of the dialog agents are important, users' chat experience is mainly determined by their favorability of the agent after the conversation.

### 5.3 Metrics derived from the conversation

Our findings showed that the coherence and sentiment in the conversation could strongly predict the chat experience. These metrics are derived from the messages and can be used by an adaptive dialog system to predict users' real-time experience and adjust the conversation through prompts. For example, an adaptive dialog system can actively evaluate the coherence and sentiment of the conversation and provide more favorable content when sentiment drops.

### 5.4 The effect of different Prompts

Users' perceived chat experience in NLG dialog system is strongly influenced by the prompt used to generate the reply. Users may prefer dialog agents with characteristics of Extroversion, Non-Neuroticism, Openness, Conscientiousness, and Agreeableness.

### 5.5 Similarity between multiple replies

Our findings showed that when presenting multiple replies to users, the similarity between agents' utterances can positively predict users' chat experience for agents with low chat experience scores. However, for agents with high chat experience scores, similarity does not have an impact on chat experience. That is, if agents are designed with bad prompts but can reply to something similar to others, compared to replying something unique but inappropriate, users might have a better impression on them.

Moreover, the results contradict with previous research [2], which states users might feel bored or confused if they receive similar replies from multiple agents. Our results indicated higher similarity between replies are positively related to or unrelated to chat experience. This may be because we limited users to chat only ten turns per round, which is not enough to get them overwhelmed with too many similar messages. Furthermore, we think designing multiple agents with different prompts so they can provide diverse replies might decrease confusion and boring level for users.

## 6 CONCLUSION

In this paper, we proposed a conceptual model to predict the chat experience in natural language generation dialog systems which obtained an R-square ($R^2$) with 0.541 in PLS-SEM. We find that users' favorability and coherence, sentiment, and similarity in the conversation are strong and positive predictors to chat experience. Moreover, our results also suggest that the design of prompts is critical in determining users' chat experience. Especially, users might prefer prompts demonstrating personalities with Extroversion, Non-Neuroticism, Openness, Conscientiousness, and Agreeableness. Through our research, an adaptive dialog system might use collected data to infer factors in our conceptual model, predict the chat experience for users through these factors, and optimize it by adjusting prompts.

## ACKNOWLEDGMENTS

We are grateful for the suggestions provided by Yuen-Hsien Tseng, Tsung-Ren (Tren) Huang, and Liang, Jyh-Chong. The study was approved by the Institutional Review Board of National Taiwan Normal University with protocol code 202206HS024. This work was supported by the Ministry of Science and Technology in Taiwan (R.O.C.) under Grants 110-2813-C-003-033-E, 109-2410-H-003-123-MY3, and 111-2634-F-002-004.